\documentclass[conference]{IEEEtran}
\IEEEoverridecommandlockouts
\usepackage{cite}
\newcommand{\email}[1]{\texttt{\href{mailto:#1}{#1}}}

\usepackage{amsmath,amssymb,amsfonts}
\usepackage{multicol,multirow,booktabs}
\usepackage{algorithmic}
\usepackage{arydshln}
\usepackage{bbding}
\usepackage{graphicx}
\usepackage{textcomp}
\usepackage{enumitem}
\usepackage{bm}
\usepackage{url}
\usepackage{threeparttable}
\usepackage{caption}
\usepackage{tabularray}
\usepackage{xcolor,hyperref}
\def\BibTeX{{\rm B\kern-.05em{\sc i\kern-.025em b}\kern-.08em
    T\kern-.1667em\lower.7ex\hbox{E}\kern-.125emX}}
\begin{document}

\title{\huge Dynamic Language Group-based MoE: Enhancing Code-Switching Speech Recognition with Hierarchical Routing}

\author{
\IEEEauthorblockN{
    Hukai Huang\textsuperscript{1,3}, Shenghui Lu\textsuperscript{1},Yahui Shan\textsuperscript{3},He Qu\textsuperscript{3},Fengrun Zhang\textsuperscript{3},Wenhao Guan\textsuperscript{1},Qingyang Hong\textsuperscript{1*},
    Lin Li\textsuperscript{2*}\thanks{*Corresponding authors.
    This work was supported in part by the National Natural Science Foundation of China under Grants 62276220 and 62371407.}
}
\IEEEauthorblockA{
\textsuperscript{1}School of Informatics, Xiamen University, Xiamen, China \\
 $^2$School of Electronic Science and Engineering, Xiamen University, Xiamen, China  \\
$^3$Kuaishou Technology Co., Ltd, Beijing, China \\ 
\email{\{qyhong,lilin\}@xmu.edu.cn}
}
}
\maketitle

\begin{abstract}
The Mixture of Experts (MoE) model is a promising approach for handling code-switching speech recognition (CS-ASR) tasks. However, the existing CS-ASR work on MoE has yet to leverage the advantages of MoE's parameter scaling ability fully. This work proposes DLG-MoE, a Dynamic Language Group-based MoE, which can effectively handle the CS-ASR task and leverage the advantages of parameter scaling. DLG-MoE operates based on a hierarchical routing mechanism. First, the language router explicitly models the language attribute and dispatches the representations to the corresponding language expert groups. Subsequently, the unsupervised router within each language group implicitly models attributes beyond language and coordinates expert routing and collaboration. DLG-MoE outperforms the existing MoE methods on CS-ASR tasks while demonstrating great flexibility. It supports different top-$k$ inference and streaming capabilities and can also prune the model parameters flexibly to obtain a monolingual sub-model. 
\end{abstract}

\begin{IEEEkeywords}
mixture of experts, code-switching, hierarchical routing, dynamic language group.
\end{IEEEkeywords}
\section{Introduction}
The mixture of experts (MoE) model can scale by incorporating multiple experts to increase model capacity while maintaining constant inference efficiency and has been extensively studied across various fields \cite{switch-transformer, outrageously, Gshard, u2++MoE, Speechmoe, kwai,SLT2024}. MoE is naturally well-suited to tackle multilingual tasks, and many studies \cite{google-smoe, MIE, MoLE} have highlighted its potential in multilingual automatic speech recognition (ASR). However, these methods still struggle to handle code-switching speech recognition (CS-ASR) tasks effectively. CS refers to the use of different languages within an utterance. However, due to language confusion\cite{bias1, bias2, degrade, youxiaojianmo1, adapter, boundary}, this presents a significant challenge. In response, the bi-encoder methods \cite{Biencoder-MoE} equip each language with its own encoder. Additionally, some methods employ language masking strategies to enhance language modeling capabilities \cite{LSCA, LAE,efficientLAE, BA-moe}. However, these methods struggle to achieve efficient inference. LSR-MoE \cite{LSR-MOE} and LR-MoE \cite{FLR-MOE} introduce two experts in each MoE layer to tackle the CS-ASR task, each dedicated to handling Chinese or English separately, and activate only one expert at a time during training and inference for computational efficiency. Although the aforementioned MoE methods have made notable progress, there is still much room for improvement. For example, we can further harness the benefits of MoE parameter scaling. Additionally, we should explore more efficient and flexible routing mechanisms and expert layer designs, which offer effective and fine-grained modeling of code-switching.
\begin{figure*}[h]
\centering
  \includegraphics[width=0.9\linewidth]{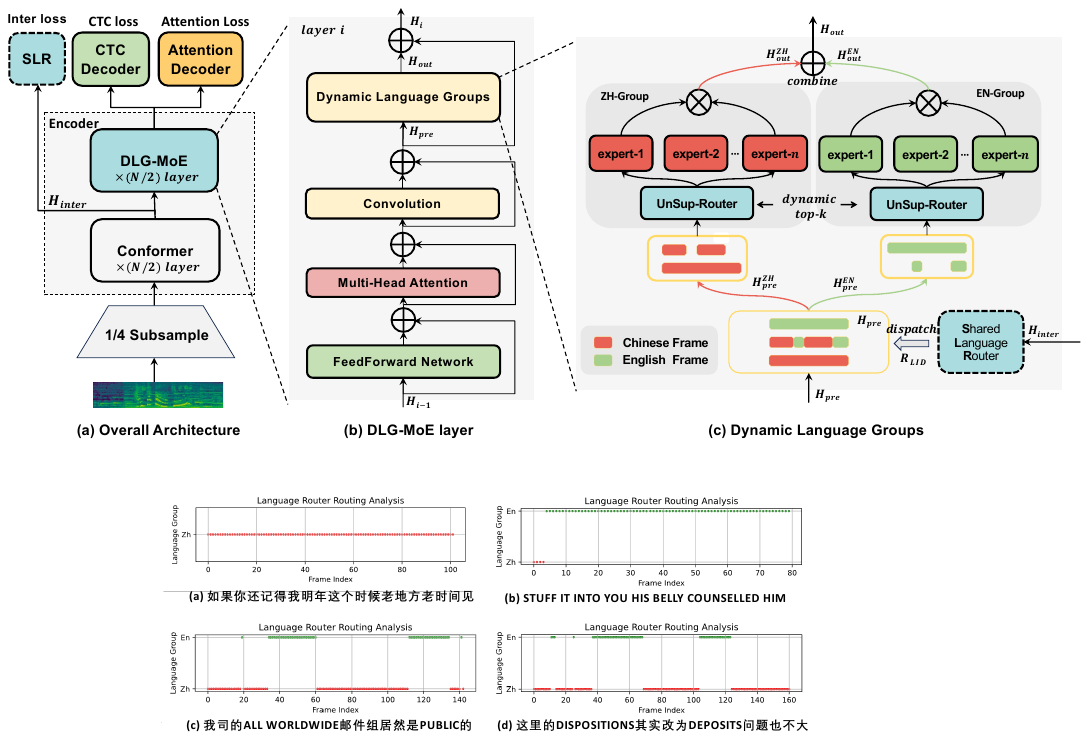} 
  \caption{\textbf{(a)} Overview of the proposed method, SLR means Shared Language Router. \textbf{(b)} DLG-MoE layer. \textbf{(c)} Dynamic Language Groups, the SLR performs language routing, followed by further routing within the language group by the UnSup-Router.}
  \label{fig:example}
\end{figure*}

The proposed DLG-MoE in this work uses hierarchical routing for efficient and fine-grained CS modeling and can scale the number of experts via language groups. Inspired by Speech MoE's \cite{Speechmoe} success with accents and domains in Chinese, we believe that different experts in MoE can implicitly model different accents or domains. Therefore, this paper proposes a hierarchical routing approach that can model different languages while also focusing on various accents and domains within the same language. Our hierarchical routing mechanism primarily relies on the language router and unsupervised router. The language router focuses on language-based routing, distributing inputs to corresponding language groups and capturing language-specific representations. The unsupervised router within the language group then further focuses on implicitly modeling accents and domains beyond language. We also introduced a dynamic top-$k$ training strategy, which supports different top-$k$ values during inference, providing a trade-off between efficiency and performance. Our contributions are summarized as follows:
\begin{itemize}[left=2pt, itemsep=2pt, topsep=1pt, partopsep=0pt, parsep=0pt]
\item The proposed hierarchical routing mechanism and dynamic language groups in DLG-MoE outperform existing methods on CS-ASR tasks. The \href{https://github.com/kaihuhuang/Language-Group}{\texttt{\textbf{code}}}\footnote{\href{https://github.com/kaihuhuang/Language-Group}{\texttt{https://github.com/kaihuhuang/Language-Group}}\label{fn:code}} and training logs have been released for reproducibility research.
\item Compared to previous related work on CS-ASR tasks, our approach demonstrates superior scalability in terms of expert capacity and supports flexible top-$k$ inference as well as streaming recognition capabilities.
\item  This work thoroughly examines the impact of different top-$k$ training strategies and the number of experts on model performance. Furthermore, we have validated the effectiveness of our methods on larger datasets and with more model parameters.
\end{itemize}

\section{Proposed Method}
\subsection{Model Architecture}
The model's overall structure is shown in Fig.1(a), which includes a CTC decoder and an attention decoder. The encoder consists of $N$ layers stacked from $N$/2 Conformer \cite{Conformer} layers and $N$/2 DLG-MoE layers. Fig1.(b) shows the DLG-MoE layer's structure, where the Conformer layer's second feedforward network is replaced with dynamic language groups. Each language group contains multiple experts and an unsupervised router (UnSup-Router). Additionally, each DLG-MoE layer shares a language router (SLR) for selecting the appropriate language group, as shown in Fig1.(c). SLR and UnSup-Router work together to perform hierarchical routing, precisely modeling different domains for different languages.

\subsection{Shared Language Router}
SLR explicitly learns the language identification (LID) task through Connectionist Temporal Classification (CTC) loss \cite{ctc}, enabling frame-level routing capabilities. Moreover, we enhance the SLR's language routing capability through a multi-task training approach that integrates ASR and LID tasks; the training loss $\mathcal{L}_{\text{inter}}$ is expressed by Equation\ref{equ:1}.
\begin{equation}
\begin{split}
\label{equ:1}
\mathcal{L}_{\text{inter}} = & -\left(\log \mathrm{P}_{\mathrm{CTC}}\left(\mathcal{Y}_{\mathrm{LID}} \mid \bm{H}_{\text{inter}} \mathbf{W}_{\text{LID}}^{D \times \mathrm{L}+1}\right) \right. \\
& \left. + \log \mathrm{P}_{\mathrm{CTC}}\left(\mathcal{Y}_{\mathrm{ASR}} \mid \bm{H}_{\text{inter}} \mathbf{W}_{\text{ASR}}^{D \times V}\right)\right)
\end{split}
\end{equation}
Here, $\bm{H}_{\text{inter}}$ represents the intermediate representation extracted by the Conformer block. $\mathcal{Y}_{\text{ASR}}$ denotes the ground truth for the ASR task, while $\mathcal{Y}_{\text{LID}}$ is derived by mapping the tokens in $\mathcal{Y}_{\text{ASR}}$ to their corresponding language IDs. The $\mathbf{W}_{\text{LID}}^{D \times \mathrm{L}+1}$ and $\mathbf{W}_{\text{ASR}}^{D \times V}$ represent the weights of the linear layer, where $D$ is the embedding dimension, $V$ is the vocabulary size, and $L$+1 represents the number of language categories plus the CTC blank token. The key point is that frame-level LID labels are not required, as the CTC can align the frames to the LID label sequence $\mathcal{Y}_{\text{LID}}$ automatically.

Since the CTC optimization objective is to maximize the sum of probabilities over all valid alignment paths, there will still be preferences among other categories, even if a particular frame has the highest probability of being a $blank$. We leverage this to smooth the CTC peak outputs and obtain frame-level LID results $\bm{R_{\text{LID}}}$. Specifically, We select the index with the highest probability among the categories other than $blank$ in the probability distribution as the language ID. This simple and causal strategy,\textbf{ which does not depend on other time steps, helps support streaming recognition.} The input $\bm{H_{\text{pre}}}$ will be decoupled into $\bm{H_{\text{pre}}^{\text{ZH}}}$ and $\bm{H_{\text{pre}}^{\text{EN}}}$ based on the $\bm{R_{\text{LID}}}$ provided by SLR and then dispatched to the corresponding language groups to extract language-specific representations.

\subsection{Dynamic Language Groups}
After determining the language attributes, the UnSup-Router within the language groups will implicitly learn other attributes and perform secondary routing. This allows experts within a language group to perform fine-grained modeling of different accents and domains within the same language. For each time step $\bm{h_{\text{pre}}^{\text{ZH}}}$ in $\bm{H_{\text{pre}}^{\text{ZH}}}$, the UnSup-Router will select the top-$k$ experts with the highest probabilities from $n$ experts to collaborate. It then combines the representations of these experts according to the learned weights to produce the output $\bm{h_{\text{out}}^{\text{ZH}}}$ for each time step, as described below:
\begin{equation}
\bm{h_{\text{out}}^{\text{ZH}}}=\sum_{i=1}^k \mathbf{w}_i \bm{e_i}
\end{equation}
\begin{equation}
\mathbf{w}=\operatorname{Softmax}\left(\operatorname{Top-k}\left(\bm{h_{\text{pre}}^{\text{ZH}}} \mathbf{W}_R^{D \times n} \right)\right)
\end{equation}
Here, \bm{$e$} and $\mathbf{w}$ denote the outputs from the selected $k$ experts and their corresponding weights. $\mathbf{W}_R^{D \times n}$ represents the weights of the linear layer in the UnSup-Router. For each time step, the input will be routed to the most appropriate $k$ experts for processing, which are then combined in a weighted manner to produce $\bm{H}_{\text{out}}^{\text{ZH}}$. Consequently, the ZH-Group and EN-Group extract language-specific representations $\bm{H}_{\text{out}}^{\text{ZH}}$ and $\bm{H}_{\text{out}}^{\text{EN}}$, respectively, combine them to obtain final output $\bm{H}_{\text{out}}$, and pass it to the next DLG-MoE layer. To support various top-$k$ inferences, we employ a dynamic top-$k$ strategy within the language groups, where the UnSup-Router randomly selects the values of top-$k$ during training. The model can be trained from scratch \textbf{without any pre-training}. All losses are jointly optimized, as illustrated below:
\begin{equation}
\mathcal{L}_{\text {total }} = \lambda_{\text{ctc}} \mathcal{L}_{\text {ctc }} + (1 - \lambda_{\text{ctc}}) \mathcal{L}_{\text {att}} + \lambda_{\text {inter}} \mathcal{L}_{\text {inter}}
\end{equation}
where $\mathcal{L}_{\text {ctc }}$ and $\mathcal{L}_{\text {att}}$ represent the CTC loss and attention loss, respectively, $\lambda_{\text{ctc}}$ and  $\lambda_{\text {inter}}$ are hyperparameters.

\section{Experiments Setup}
\subsection{Datasets}
Our experiments are conducted on the ASRU-2019 Mandarin-English code-switching challenge dataset \cite{asru}, along with a subset of 460 hours from Librispeech \cite{librispeech}. In the ASRU-2019 challenge dataset, the English has a Chinese accent, while Librispeech is standard. The total duration of the datasets used in our experiments is approximately \textbf{1100 hours}, \textbf{consistent with previous works} \cite{Biencoder-MoE, LSR-MOE, FLR-MOE}. We use Librispeech-clean, ASRU-Man-test, and ASRU-CS-test datasets for the test set. Additionally, we have validated our method on in-house multi-accent, multi-domain industrial data, which, combined with the ASRU-2019 challenge dataset, totals \textbf{6,000 hours} of training data, and evaluate the performance of the English and Chinese parts in the ASRU-CS-test.
\subsection{Model Configurations}
We conducted experiments on the WeNet \cite{wenet} toolkit, utilizing its U2++ framework. For acoustic features, 80-dimensional log-mel filter banks (FBank) are extracted with a step size of 10ms and a window size of 25ms. All our experimental configurations used a 6-layer Transformer \cite{speech-transformer} decoder. We set $d^{model}$=256, $n^{heads}$=4 and $d^{ffn}$=2048. Causal convolution was used during training to support streaming. We built \textbf{Baseline-12} and \textbf{Baseline-18} with 12 and 18 Conformer layers, respectively. All MoE models had 12 layers, with the first 6 being vanilla Conformer layers and the last 6 incorporating MoE layers. We constructed the following original MoE models with different numbers of experts: \textbf{Sparse-MoE-2e}, \textbf{Sparse-MoE-4e} and \textbf{Dense-MoE-4e}. We also built the \textbf{DLG-MoE-2e}, \textbf{DLG-MoE-4e}, and \textbf{DLG-MoE-8e}, where the total number of experts is evenly divided among the languages (It can be designed flexibly). For the experiments verified on 6000 hours of data, we used $d^{model}$=512 and $n^{heads}$=8 for Baseline-18-\textbf{L} and DLG-MoE-8e-\textbf{L}. 
% \XSolidBrush
\begin{table*}[h]
% \small
\footnotesize
\label{tab:main}
\centering
\caption{\small{The results (\%) of various systems on Libri-clean(EN), ASRU-Man(ZH), and ASRU-CS(CS) test sets. Inference size and FLOPs reflect the computational complexity of the model. The symbol ‘\textbf{*}’ indicates that the top-k during inference does not match the training.}}
\setlength{\tabcolsep}{9pt} % 设置列之间的空白宽度
\renewcommand{\arraystretch}{0.9} % 设置行高
\begin{tabular}{clcclllll} 
\toprule
\textbf{System}               & \textbf{Model }                                                      & \textbf{Top-k Train}  & \textbf{Size}                   & \textbf{Inference Size}    & \textbf{FLOPs↓} & \textbf{EN(\%)↓} & \textbf{ZH(\%)↓} & \textbf{CS(\%)↓}  \\ 
\hline
$S1$                   & Baseline-12                                                 & top-1                     & 51M                    & 51M(top-1 decode) & 24.8G   & 6.89    & 3.35    & 9.93     \\
$S2$                   & Baseline-18                                                & top-1                     & 72M                    & 72M(top-1 decode) & 36.2G   & 6.52    & 2.93    & 9.52     \\
$S3$                   & Dense-MoE-4e                                                & top-4                     & 72M                    & 72M(top-4 decode) & 34.3G   & 6.09    & 2.73    & 9.36     \\

$S4$                   & Sparse-MoE-4e                                               & top-2                     & 72M                    & 60M(top-2 decode) & 27.8G   & 5.99    & 2.70    & 9.38     \\
$S5$                   & Sparse-MoE-4e                                               & dynamic                  & 72M                    & 60M(top-2 decode) & 27.8G   & 6.12    & 2.89    & 9.62     \\ 
$S6$                   & Sparse-MoE-2e                                               & top-1                     & 60M                    & 51M(top-1 decode) & 25.0G   & 6.24    & 2.91    & 9.45     \\
$S7$                   & DLG-MoE-2e                                               & top-1                     & 60M                    & 51M(top-1 decode) & 25.0G   & 5.72    & 2.66    & 9.27     \\  
\hline\hline
\multirow{2}{*}{$S8$}  & \multirow{2}{*}{DLG-MoE-4e}                                 & \multirow{2}{*}{top-2}    & \multirow{2}{*}{72M~}  & 51M(top-1 decode)\textsuperscript{\textbf{*}} & 25.0G   & 5.39    & 2.77    & 9.81     \\
                     &                                                             &                          &                        & 60M(top-2 decode) & 27.8G   & 5.31    & 2.55    & 9.09     \\ 
\hdashline[1pt/1pt]
$S9$                   & \begin{tabular}[c]{@{}l@{}}DLG-MoE-4e\\(w/o UnSup-Router)\end{tabular} & top-2                     & 72M                    & 60M(top-2 decode) & 27.8G   & 5.44    & 2.67    & 9.23     \\ 
\hdashline[1pt/1pt]
\multirow{2}{*}{$S10$}  & \multirow{2}{*}{DLG-MoE-4e}                                 & \multirow{2}{*}{dynamic} & \multirow{2}{*}{72M}   & 51M(top-1 decode) & 25.0G   & 5.51    & 2.54    & 9.16     \\
                     &                                                             &                          &                        & 60M(top-2 decode) & 27.8G   & 5.42    & 2.52    & 9.10     \\ 
\hline\hline
\multirow{2}{*}{$S11$}  & \multirow{2}{*}{DLG-MoE-8e}                                 & \multirow{2}{*}{top-1}    & \multirow{2}{*}{103M}  & 51M(top-1 decode) & 25.0G   & 5.62    & 2.58    & 9.30     \\
                     &                                                             &                          &                        & 60M(top-2 decode)\textsuperscript{\textbf{*}} & 27.8G   & 5.82    & 2.66    & 9.52     \\ 
\hdashline[1pt/1pt]
\multirow{2}{*}{$S12$} & \multirow{2}{*}{DLG-MoE-8e}                                 & \multirow{2}{*}{top-2}    & \multirow{2}{*}{103M~} & 51M(top-1 decode)\textsuperscript{\textbf{*}} & 25.0G   & 5.84    & 2.64    & 9.77     \\
                     &                                                             &                          &                        & 60M(top-2 decode) & 27.8G   & \textbf{5.26}    & \textbf{2.39}    & 9.10     \\ 
\hdashline[1pt/1pt]
\multirow{2}{*}{$S13$} & \multirow{2}{*}{DLG-MoE-8e}                                 & \multirow{2}{*}{dynamic} & \multirow{2}{*}{103M~} & 51M(top-1 decode) & 25.0G   & 5.36    & 2.45    & 9.05     \\
                     &                                                             &                          &                        & 60M(top-2 decode) & 27.8G   & 5.29    & 2.42    & \textbf{8.92}     \\
\bottomrule
\end{tabular}
\end{table*}
$\lambda_{\text{ctc}}$ and $\lambda_{\text{inter}}$ were set to 0.3 and 0.1, respectively. All models are trained with the Adam optimizer for \textbf{50 epochs} with $25,000$ warmup steps, and the peak learning rate is $0.001$. More detailed configuration is available in the provided link\textsuperscript{\ref{fn:code}}.
\subsection{Evaluation Metrics}
We use character error rate (CER), word error rate (WER), and mixed error rate (MER) to evaluate the model performance on the \textbf{ASRU-Man(ZH)}, \textbf{Librispeech-clean(EN)} and \textbf{ASRU-CS(CS)} test sets, respectively. Floating-point operations (FLOPs) measured on 20-second input audio are used to measure the computational complexity of the model.
\section{Experiments Results}
\subsection{Analysis of Results for Different MoE Structures}
Table I shows the performance of different MoE structures on the monolingual (EN, ZH)  and code-switching (CS) test sets, from system $S1$ to $S9$; we have the following findings:
\begin{itemize}[left=0pt, itemsep=1pt, topsep=0.5pt, partopsep=0pt, parsep=0pt]
    \item Widening the model with MoE ($S3$ $vs$ $S2$) is more effective than deepening the model ($S2$ $vs$ $S1$). Additionally, the Sparse-MoE selectively activating specific experts results in better overall performance and computational efficiency than activating all experts in Dense-MoE ($S4$ $vs$ $S3$).
    \item The hierarchical routing-based DLG-MoE shows advantages over Sparse-MoE ($S7$ $vs$ $S6$, $S8$ $vs$ $S4$) due to the language router explicitly decoupling representations of different languages, allowing each language group to focus on modeling its language and maintaining language-specific solid capabilities.
    \item The expert weights learned by the UnSup-Router within the language groups are also important, showing advantages over a system with equal expert weights and no UnSup-Router ($S8$ $vs$ $S9$). This indicates that the UnSup-Router can further model attributes beyond language. The ablation study composed of $S4$, $S9$, and $S8$ demonstrates the effectiveness of hierarchical routing($S4$$\rightarrow$$S9$$\rightarrow$$S8$).
\end{itemize}
Since previous MoE works lacked open-source code, we present results from the original paper, including Biencoder-MoE \textbf{(9.76\%)}\cite{Biencoder-MoE}, LSR-MoE \textbf{(10.30\%)}\cite{LSR-MOE}, and FLR-MoE \textbf{(9.70\%)}\cite{FLR-MOE}. Our method demonstrates better performance.

\subsection{Analysis of Expert Capacities and Top-k strategy}
Table I systematically presents the impact of different top-$k$ training strategies and the number of experts on the performance of DLG-MoE for systems $S7$ to $S13$. The experimental results can be summarized as follows:
\begin{itemize}[left=0pt, itemsep=0.5pt, topsep=0.5pt, partopsep=0pt, parsep=0pt]
    \item DLG-MoE can scale to more experts and achieve better results ($S7$, $S10$, $S13$), and top-2 training generally outperforms top-1 training ($S12$ $vs$ $S11$). Extending experts in Sparse-MoE yields limited benefits ($S4$ $vs$ $S6$), likely because mismatches between the number of experts and languages make implicit language modeling more difficult for the router, a conclusion similar to LSR-MoE \cite{LSR-MOE}. The design of language groups in DLG avoids this issue, leading to more noticeable improvements with scaling.
    \item The dynamic top-$k$ training strategy enables the model to perform well across different top-$k$ values during inference ($S10$, $S13$) while providing a trade-off between performance and efficiency. For systems with a fixed top-$k$ strategy ($S8$, $S11$, $S12$), mismatched top-$k$ values between training and inference can cause significant performance drops. We also applied this dynamic strategy to Sparse-MoE but found worse results ($S4$ $vs$ $S5$), likely because dynamic top-$k$ makes implicit language modeling more challenging in Sparse-MoE. Notably, $S13$ with dynamic top-$k$ even outperformed the fixed top-1 and top-2 systems ($S11$, $S12$), demonstrating that dynamic top-$k$ enhances model generalization, especially when there are more experts.
\end{itemize}

Our system $S13$ maintains the same number of activated parameters and similar FLOPs as baseline $S1$ in top-1 decoding, yet achieves up to \textbf{22.2\%}, \textbf{26.9\%}, and \textbf{8.9\%} relative improvement in EN, ZH, and CS, respectively. Our method also demonstrates substantial advantages with larger datasets and model parameters, as shown in Table \ref{tab:large}.

% 22.2% 26.9% 10.2%
% \usepackage{tabularray}
\begin{table}
\centering
% \small
\footnotesize
\setlength{\tabcolsep}{4pt} % 设置列之间的空白宽度
\caption{\small{Results (\%) with larger datasets and parameters.}}
\label{tab:large}
\renewcommand{\arraystretch}{0.85} % 设置行高
\scalebox{0.92}{  % 在此处添加缩放指令
\begin{tblr}{
  cell{1}{1} = {r=2}{},
  cell{1}{2} = {r=2}{},
  cell{1}{3} = {r=2}{},
  cell{1}{4} = {c=3}{c},
  cell{4}{1} = {r=2}{},
  cell{4}{2} = {r=2}{},
  hline{1,6} = {-}{0.08em},
  hline{2} = {4-6}{},
  hline{3} = {1-3}{},
  hline{3} = {4-6}{0.03em},
  hline{4} = {-}{dashed},
}
\textbf{Model} & \textbf{Size} & \textbf{Inf. Size} & \textbf{ASRU-CS} &      &       \\
               &               &                    & MER              & CER  & WER   \\
Baseline-18-L    & 173M          & 173M(top-1)         & 6.65             & 4.50 & 24.25 \\
DLG-MoE-8e-L     & 228M          & 100M(top-1)         & 6.16             & 4.12 & 22.88 \\
               &               & 118M(top-2)         & \textbf{6.11}             & \textbf{4.08} & \textbf{22.78} 
\end{tblr}
}
\end{table}

\begin{table}
\label{tab:streaming}
\centering
% \small
\footnotesize
\caption{\small{Streaming results (\%) for different chunk sizes.}}
\label{tab:example}
\setlength{\tabcolsep}{2pt} % 设置列之间的空白宽度行高
\renewcommand{\arraystretch}{0.9} % 设置行高
\begin{tabular}{ccccccc} % 定义五列
\toprule
\textbf{Mode} & \textbf{Size} & \textbf{Specify Lang.} & \textbf{EN} & \textbf{ZH} & \textbf{CS} & \textbf{LID↑} \\
\midrule
NonStreaming & 103M & \XSolidBrush & 5.29 & 2.42 & 8.92 & 99.4 \\
chunk=16(640ms) & 103M & \XSolidBrush & 6.21 & 2.82 & 9.89 & 99.1 \\
chunk=8(320ms) & 103M & \XSolidBrush & 6.69 & 3.02 & 10.31 & 99.1 \\
chunk=8(320ms) & 72M & \CheckmarkBold & \textbf{6.64} & \textbf{2.98} & - & - \\
\bottomrule
\end{tabular}
\end{table}
\subsection{Streaming Capability and Flexibility}
This section demonstrates the additional advantages of our proposed method. As shown in Table \ref{tab:example}, the model can still maintain reasonable performance in streaming scenarios with different chunk sizes due to the frame-synchronous LID decoding supported by the language routing. Notably, as shown in the last row of Table \ref{tab:example}, in the monolingual scenario, we can even specify the language for inference and prune the parameters of the other language groups to obtain a monolingual sub-model (103M$\rightarrow$72M). Alternatively, each time step can be directly routed to a specific language group, thereby avoiding additional language routing overhead and potential misrouting. The results also show that ensuring the correct routing of all frames provides a slight improvement, which indirectly confirms that the language routing was already accurate enough. The token-level LID results shown in Table \ref{tab:example}, with over 99\% accuracy, further confirm the accuracy of the language routing. Fig.\ref{fig:example} shows the visualization of the frame-level routing results of the language router in $S13$. The language router has high routing accuracy, especially for monolingual utterances. It also tends to dispatch the first few frames of English to the ZH-Group, which may correspond to frames with no speech. For CS utterances, a high routing accuracy was also observed. 
\begin{figure}[h]
\centering % 如果需要居中显示图片，可以取消注释这行
  \includegraphics[width=1.0\linewidth]{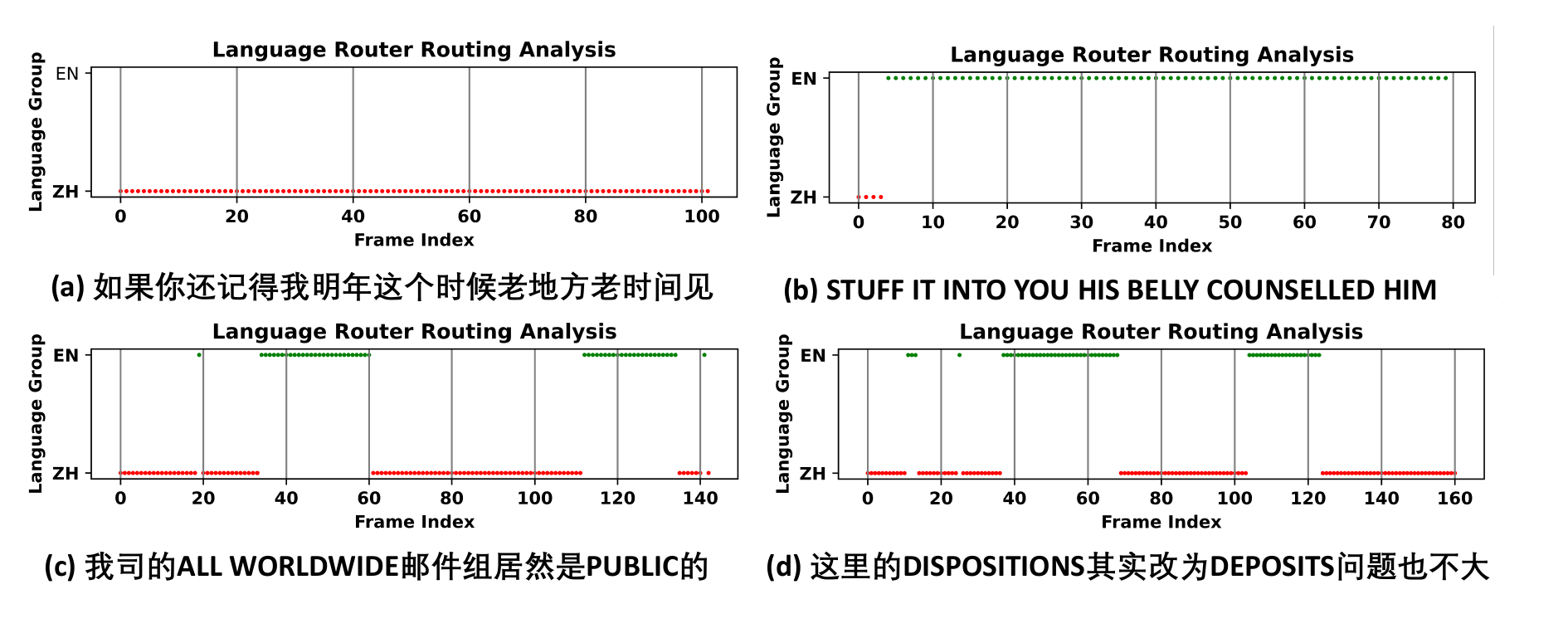} % 替换为你的图片文件名
  \caption{\small{Visualisation of SLR's frame-level routing results, where red represents Chinese and green represents English.}}
  \label{fig:example}
\end{figure}
\section{Conclusions and future work}
This paper introduces a novel MoE model that is efficient and flexible for code-switching tasks. The DLG-MoE model excels with its hierarchical routing and dynamic language groups, delivering superior performance while maintaining comparable computational costs to the baseline. The study further investigates the impact of expert capacity and different top-$k$ training strategies on model performance, demonstrating that the language groups and the proposed dynamic top-$k$ training strategy can fully leverage the advantages of MoE, better scaling the number of experts to handle code-switching tasks. In the future, we plan to explore the potential of hierarchical routing across a broader spectrum of languages, accents, and domains.

\bibliographystyle{IEEEtran}
\bibliography{refs}
\end{document}